# Mining atmospheric data

Chaabane Djeraba
*Univ. Lille, CNRS, Centrale Lille, UMR 9189 CRIStAL,*
IRCICA, 50, avenue Halley, 59650 Villeneuve d'Ascq, France
chabane.djeraba@univ-lille.fr

Jérôme Riedi
*ICARE, LOA, University of Lille*
Bat. M3, Cité Scientifique, Villeneuve d'Ascq, France
jerome.riedi@univ-lille.fr

*Abstract*— This paper overviews two interdependent issues important for mining remote sensing data (e.g. images) obtained from atmospheric monitoring missions. The first issue relates the building new public datasets and benchmarks, which are hot priority of the remote sensing community. The second issue is the investigation of deep learning methodologies for atmospheric data classification based on vast amount of data without annotations and with localized annotated data provided by sparse observing networks at the surface. The targeted application is air quality assessment and prediction. Air quality is defined as the pollution level linked with several atmospheric constituents such as gases and aerosols. There are dependency relationships between the bad air quality, caused by air pollution, and the public health. The target application is the development of a fast prediction model for local and regional air quality assessment and tracking. The results of mining data will have significant implication for citizen and decision makers by providing a fast prediction and reliable air quality monitoring system able to cover the local and regional scale through intelligent extrapolation of sparse ground-based in situ measurement networks.

## I. Introduction

Pattern recognition in atmospheric data is a very active research area. Pattern recognition is a great research area in machine leaning and computer vision. Among pattern recognition techniques, Artificial Neural Networks, and more particularly deep learning are the most popular. The main advantage of the deep learning is the dealing with huge data sets, that may be augmented artificially based on few annotated data sets. Furthermore, the features are created automatically compared to handcrafted machine learning. We jump from sequential processes of learning to one complete process, where the deep artificial network converges input data to output annotated data. The multilayers and the back propagation contribute to these important advances in machine learning. Deep learning is becoming suitable for several applications, such as air quality and pollution prediction. In the literature, several comparisons of machine learning methods on atmospheric data, including satellite images have been investigated [2]. Their evaluations demonstrate clearly [3] that in general deep learning presents better performances compared to handcrafted learning. In contrast to handcrafted learning, deep learning automatically infers suitable features for pattern recognition, from atmospheric data. However, deep learning requires significant amount of ground truth data, and the augmentation technique is subject to investigations and open research area in the atmospheric domain.

The central problem is huge quantity of satellite images without annotations and with localized annotations provided by sparse observing networks at the surface. The problem is dealing with huge data, without or with limited ground truth. Therefore, the development of algorithms and the acquisition approaches for the ground truth is an important machine learning step. Then, the next step would be the development of deep learning architectures, destined to process high dimension data, 100s to 1000s dimensions satellite images. Meanwhile, new methods of data transfer learning and domain adaptation algorithm should be investigated to consider the existing satellite image databases [4]. There are several applications of pattern recognition in atmospheric data. One of the most popular is air pollution detection and tracking.

The scope of the paper is the discussion of scientific challenges which are hot priorities of the atmospheric scientific community:

- Understanding of complex atmospheric phenomena.
- Dealing with big data and high dimensions of hyperspectral images.
- Building new public datasets and benchmarks.
- Developing new machine learning approaches for efficient data classification of big data (without annotations) localized and limited annotated data provided by sparse observing networks at the surface.
- Modelling and predicting atmospheric phenomena, such as the air quality and the pollution level.

The paper is composed of several sections. The section 2 presents scientific issues of atmospheric data mining. The section 3 presents the societal context, including social, economic impact. And the final section is the conclusion.

## II. Technical Issues

### A. Understanding of complex atmospheric phenomena

In our context of atmospheric phenomena understanding, the governing equations of the plume and cloud are not perfectly known. The Navier-Stokes equations and its derivatives are not valid for all plume and cloud. At present, the problem remains open. These nonlinear equations which are not simple and still not completely defined mathematically (analysis, meteorology, quality of air prediction) can generate extremely complex behaviors, such as turbulence. We can approach these phenomena, chaotic or not, from a non-equilibrium statistical physics point of view, from theoretical methods (temporal correlation functions, convolution and filtering), but it remains difficult to predict the fine behaviors of turbulence from the equations. However, most of the flows around us (Aerosols, water, air, oil...) are non-Newtonian and turbulent.

So, deal with the limitations of the governing equations of fluids, the central questions are: how machine learning interacts with equations, such as Euler equations, the Navier-Stokes equations, etc., to understand the atmospheric phenomena in motion? How machine learning contribute to understand the transition between a simple fluid behavior (laminar flow) and a behavior with eddies (atmospheric turbulent flow)? How machine learning interacts with the physics dedicated to the study of the behavior of atmospheric phenomena and the associated internal forces? How motion of atmospheric phenomena is modeled in machine learning technologies? Machine learning interacts with mathematics and continuous functions. Because, the atmospheric particles (e.g. Aerosol) are small enough to be analyzed mathematically, and large enough, compared to molecules, to be described by continuous functions.

The machine learning solution to the atmospheric dynamics understanding usually involves the calculation of various spatial-temporal plume cloud properties, such as celerity, force, intensity and thermal. Computational aerosol mechanics consists in studying the motions of the plume, or their consequences, by tackling the mathematic equations ruling the physics phenomena of plume cloud. The convergence of the results is a balance between the completeness of the equations and machine learning resources.

*B. Big atmospheric data*

Atmospheric data is typical of the big data phenomena. Several agencies responsible for the space program, such as the National Center of Space Studies (CNES) or the National Aeronautics and Space Administration (NASA), operate observing missions that generate increasingly huge quantity of atmospheric data. In particular in Europe, the Sentinel missions, part of the Copernicus program are generating several TB of data every day. We will take the example of ICARE Data and Services Center, which collaborate with French and international space agencies. ICARE interacts strongly with academic research and environment agencies, on the basis of atmospheric database indexing and retrieval by content. The databases are composed of satellites images and earth observations of atmospheric phenomena, mainly cloud, plume, sea and radiations. It provides data necessary to scientific research in atmospheric and climate sciences. ICARE belongs to an international network, composed of atmospheric data centers and international space agencies. It belongs also to the French Data Infrastructure for the Atmosphere created in 2014. ICARE is a government space agency, supported by the national center of space studies and localized in Lille University. The main activities of ICARE include collecting, processing, archiving, and disseminating data. ICARE also develops data visualization and analysis tools, implement and operate operational processing codes and provide support to atmospheric science users with dedicated human and hardware resources. The specific expertise of ICARE is the exploitation of the rich satellites data records, the combination of the various observations to derive geophysical parameters, and easy online access to products. ICARE hosts a very large computing system, located at the IT department of Lille University: 1300 referenced data sets available online (3765 TB archive), 350 data sets collected daily, 950 derived products and associated quick looks generated routinely, 90 processing codes, Incoming daily rate of 1.1 TB, total daily increase of the archive: 1.4 TB/day (ingest and production), 136 million files archived. The services and production system include 124 servers totaling more than 2400 cores. In 2019, the production system provided 18,000 cores-days of computation. ICARE web services are used by more than 3500 registered users worldwide for data access, 196 registered users for computer resources access, with 17,000 unique visitors and 43,000 visits per month.

*C. High dimension atmospheric data*

Generally, in the computer vision, image pixels are coded within format such as RGB, RGB-D, 3D models, greyscale, etc. The red, green and blue values are represented respectively by three wavelength measurements, situated between 400 and 700 nano-meters. However, atmospheric data are generally multispectral images. So the pixel of the image is coded in more than four wavelengths and less than 10, with applications on geology, atmosphere and agriculture. Multispectral images contain more information than RGB images. Hyperspectral images consider 11 to several hundred of wavelengths, with application on remote sensing, geoscience and environment. The hyperspectral image [6] is composed of series of images acquired at different spectrum, from which can be determined geophysical parameters, such as aerosol index, aerosol optical depth or other atmospheric properties. The process of inferring atmospheric properties from hyperspectral measurement is usually computationally expensive and needs high indexing, retrieving and mining systems. There is therefore a critical need in atmospheric science to develop efficient technics for data analysis in order to address the challenge of handling increasingly massive datasets.

Below (Fig. 1) is an illustration of a pollution plume generated by massive wild fire in California on September 10, 2020. The true color image (right bottom insert) provided by the geostationary imaging sensor on GOES-West shows the massive smoke plume off the coast of California. The Aerosol Index product, derived from the hyperspectral imager TROPOMI on board the Sentinel-5 Precursor (large image) for the same day, is mapped along with location of the AERONET ground-based network sites. This illustrates the challenge of linking global satellite products with localized ground truth provided by sparse observing networks at the surface.

Mining, indexing and processing satellite images for understanding atmospheric phenomena is well established in the domain of earth observation. Despite the success of these systems for tackling various challenging problems, very serious challenges are still active and highlight future researches. The rich spectral information is also the source of problems for machine learning [1]. The challenge here is the curse of the dimensionality. It implies analyzing high-dimensional data. When the data dimension expands, the representation space expands so rapidly that the representation of the data become complex for the learning methods. The high dimension of the data expands the quantity of the data required for pattern recognition, indexing, retrieving and mining of atmospheric data. So, there are relationships between the high dimension of the data and the complexity of indexing, retrieving and mining.

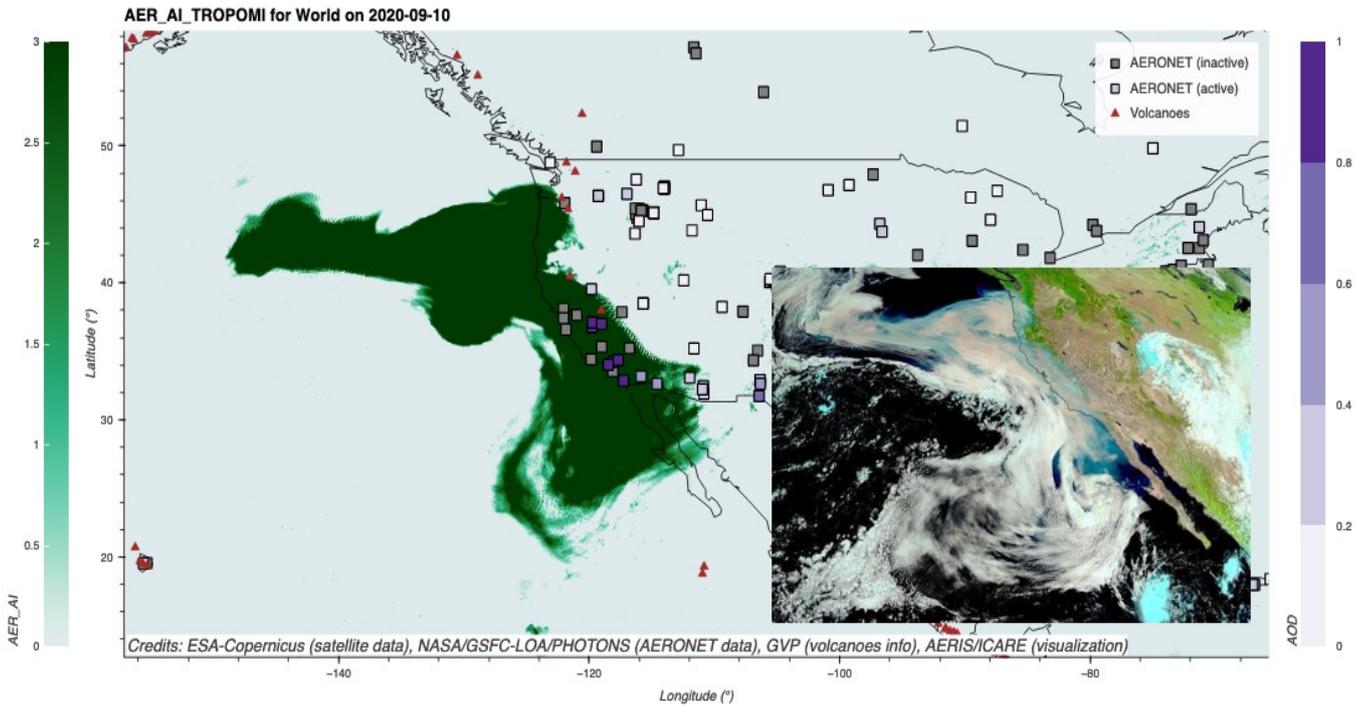

Fig. 1. Illustration of Aerosol Index retrieved from TROPOMI over western north America exhibiting a massive smoke plume from wildfire in California on September 10th, 2020.

*D. Benchmark*

Developing the benchmark is the top priority of atmospheric physics community. It consists of building new public datasets and benchmarks (datasets, ground truth data and protocol of evaluation) for several applications such as air quality prediction. The objective in our context is the construction of large-scale dataset in-the-wild. Such datasets will have deep impact in the field of machine learning in atmospheric data, similar to the creation of ImageNet [5] for the growth of computer vision and deep learning. This objective addresses the relative weakness of the presence of large-scale datasets and protocol of evaluations. It will contribute to the development of normalized approaches to evaluate concurrent algorithms. With such standard procedures, experts in atmospheric physics may select the suitable approaches for their objectives.

The necessity of building the benchmark is based on the fact that, without a set of large-scale databases and normalized approaches to evaluate concurrent algorithms in-the-wild, researchers cannot find the best solution for their problem. The scientific community used various datasets with several benchmarking protocols. However, it is practically impossible to compare two methods proposed in two different papers. Furthermore, generally, it is difficult to reproduce the experiments of the approaches, because the codes are not available. These factors decrease significantly the reproducibility of the methods and discourage the scientific community in the development of new methods. I would like to underline some agencies that develops the platform for evaluating the different machine learning. I mention IEEE Geoscience and Remote Sensing society, IEEE GRSS annual data fusion contest and Rochester Institute of Technology. From the few available public datasets, most are for land cover classification. This has encouraged the development of more methods for land cover classification than any of the other machine learning tasks. In other domains, such as target/anomaly detection, generally, researchers use simulated data or private datasets to measure the performance, because there are few public datasets for these tasks.

*E. Deep learning*

The point is to develop deep learning approaches for atmospheric data modeling, on the basis of few annotated data. Deep learning requires massive data, that is why they return bad performances, with few annotated data. To deal with this problem, several approaches may be investigated.

- The first one is new algorithms for transfer learning and domain adaptation in the context of hyperspectral image analysis.

- The second one is the combination of the residual learning and feature fusion.

- The third one is the study of the deep learning architectures.

- The fourth one concerns strategies to improve classification performance.

- The fifth one is the development of approaches with high speed and low energy consumption. Deep learning requires massive computing which is high energy consuming. We are thinking about spike neural networks which are very fast and very low energy consuming. The recent advances in the spike neural network architecture make this option possible.

The issue starts from a simple and fundamental observation. Spatial sensors produce automatically at large-scale satellite images, and this production is cheap, after the initial cost of the sensors. However, creating ground truth information for even small area is very expensive. That is why, today, we have huge quantities of hyperspectral images, without annotations or with limited annotations, provided by earth sparse observing networks. In the future, with decreasing costs of earth observing networks, more and more data will be annotated and the gap between annotated data and data without annotations will be reduced. Many investigations are still required on unsupervised deep learning to learn rich and robust prediction models.

*F. Spike Neural Network (SNN)*

Spike neural networks, very inspired by the human biological model compared to other artificial neural networks, have the property of being able to be implemented very efficiently on hardware, in order to create very fast architectures with very low energy consumption. Spike neural networks implement the rule of plasticity as a function of time of occurrence of spikes (STDP), inspired by biological learning. In addition, they are well placed to address the problem of learning with little or no labeled data. The literature around spike neural networks mainly uses simple data (e.g. MNIST, 70,000 images of handwritten digits), and the STDP rule has been little studied on more complex data, such as image sequences, with strong movements/spatio-temporal connotations, such as satellite/atmospheric images. Spike neural networks currently have lower performance than traditional deep learning methods (e.g. CNN), due to the constraints of the STDP rule: low frequencies, use of local computational memory, reduction of back propagation, reduction in the number of layers. A scientific approach consists of the extension of spike models [7], trained through plasticity as the function of the time of occurrence of spikes, to take into account the features of deep learning for the classification of satellite images. The objective is to improve the robustness of current models, while respecting the balance between the current state of neuromorphic hardware architectures and their future evolutions.

## III. SOCIETAL ISSUES

*A. Application: Air quality modeling*

A popular application of machine learning in atmospheric physics is the prediction of the air quality. Air quality is defined as the pureness indicator of the air, measured by the concentrations in the air of various atmospheric constituents (gases and aerosols). Air quality is conversely proportional to air pollution, with deep impact in public health. In big cities, air quality is unpleasant and pollution level high. The pollutants and haze are present. The measure of the air quality, versus air pollution is standardized. However the mathematic formulas are little bit different between countries. The indices have different intervals to determine the state of air pollution and its impact in public health. The most used index is the air quality index developed by the United States Environmental Protection Agency. Big cities, with their cars, buses, boats, plane and manufacturer, transmit large quantities of toxic or harmful pollutants to the atmosphere such as $SO_2$, $NO_2$, CO, PM and toxic organics. These air contaminants impact our environment, including plants, animal, people and building, and more generally, serious health problems impacting the whole world such as global warming and climate change. We can list some health problems impacted by air pollution: pulmonary, cardiology, respiratory problems by inhalation. Air quality prediction would determine future concentrations and sources of air contaminant.

The interdependent issues are in the service of developing a fast prediction model for local and regional air quality assessment and tracking based on the synergy between satellite observations of tropospheric aerosols, meteorological model and ground-based observation networks. The application challenge we want to address is to establish a link between satellite observation of aerosol properties (big data) and particulate matter (PM) concentration in the atmospheric boundary layer (limited ground truth data). For this, machine learning techniques, and more particularly the new architecture of deep learning will operate on data (space based, model, ground network) in order to provide a fast, statistically optimized model of air quality able to provide relevant information to decision makers at local regional scale. The approach we want to investigate is a hybrid approach combining statistical optimization based on deep learning (IA) and model scaling to infer Particulate Matter of size 2.5 micrometers or sizes 10 micrometers or less concentration near the surface from satellite observations performed by different sensors.

*B. Social impact*

The social impact is huge. The monitoring and forecast of air quality are essential to evaluate the negative impact of pollution on health. The World Health Organization announced that in the year 2016, the bad air quality caused 4.2 million deaths. Furthermore, Worldwide, ambient air pollution is estimated to cause about 16% of the lung cancer deaths, 25% of chronic obstructive pulmonary disease (COPD) deaths, about 17% of ischemic heart disease and stroke, and about 26% of respiratory infection deaths. Particulate matter pollution is an environmental health problem that affects people worldwide, but low- and middle-income countries disproportionately experience this burden.

In addition to direct health consequences, our ability to monitor and predict air quality is key in developing mitigation strategies both at social and economic level. Policy makers need relevant information to evaluate how public policies impact the local and regional air quality, while citizens may want to adjust their behavior to adapt to pollution level by changing their commuting hour or adapting their biking itineraries to follow roads with better air quality.

Finally developing information services on air quality that are both accurate and relevant to local population (high spatial and temporal resolution needed) might help raise awareness about this problem and need to reduce air pollution.

*C. Research support*

The paper is held in the context of the synergies of interdisciplinary academic experts (AI computing/Earth Remote Sensing), private sectors partners (SPASCIA) and public organization (ATMO-Hauts-de-France) in cooperation with international scientists from international organizations (NASA, SRON, University of Western Australia) to develop innovative approaches and services for air quality monitoring. The PHIDIAS project [8], [9] is another example of European research support. PHIDIAS is an H2020 project co-financed by the Innovation and Networks Executive Agency (INEA) under the European Union's Connecting Europe Facility (CEF). The project addresses the development and concrete realization of a set of high-performance computing based interdisciplinary services and tools to exploit large satellite datasets of public European interest provided by Satellite observation of Earth.

By focusing on satellite data provided mainly by Europe's Sentinel program, our project enhances the activity and added value of EU funded programs (Sentinel/Copernicus [10]). It is also in line with several regional projects related to climate change impact (CPER Climense [11]), and will contribute in long term to 'critical mass' required to tackle the challenges of bridging high-performance computing and artificial intelligence techniques with big data originating from earth observation systems. It also clearly responds to the "need to respond to economic and societal challenges", in particular related to air quality impact on health.

The long-term objective of the paper is to open the huge quantity of data initially acquired for environmental scientific research to other scientific research areas. These scientific researches concern the valorization of voluminous data in scientific domains such as machine learning, deep learning, big data management, energy consumption, health, medicine, transport, etc. The research is by essence interdisciplinary with computer vision/machine learning and atmospheric physic. The research we propose is also in line with the strategic development of the Data Terra [12] research infrastructure regarding the development of Artificial Intelligence applications for environmental research.

IV. CONCLUSION

Atmospheric data mining is by essence inter-disciplinary, with several challenges: managing big data, high dimension data, public benchmarks and computer vision/deep learning approach with big data and without or with very limited annotations. It is an inter-disciplinary area between computer vision/deep learning area and atmospheric physics, with deep impact on human life. A popular application is the study the problem of air quality and pollution modeling using hyperspectral images and real-time environmental data, with focus on learning in big data and high dimension data without or with few ground truths. To deal with these challenges, we aim at building public datasets and benchmarks for air quality in collaboration with the different partners, under the supervision of machine learning experts and data centers.

ACKNOWLEDGMENT

We would thank IRICICA – USR CNRS for its support. AERIS/ICARE is funded by CNES, CNRS, University of Lille and Région Hauts-de-France.